\documentclass[authoryear,preprint,review,12pt]{elsarticle} 

\usepackage{graphics} 
\usepackage{float}
\usepackage{graphicx}
\usepackage{wrapfig} 

\usepackage[draft]{todonotes}   % notes shown

\usepackage[labelfont=bf]{caption}
\usepackage{caption}
\usepackage{color}
\usepackage{amsmath} % Mathe
\usepackage{mathtools} % Mathe
\usepackage{amsfonts} % Mathesymbole
\usepackage{amssymb}
\usepackage{array,longtable}

\usepackage{array}
\usepackage{booktabs}
\usepackage{rotating}
\usepackage{multirow}
\usepackage{multicol}
\usepackage{lscape}
\usepackage{subfig}

\usepackage[utf8]{inputenc}
\usepackage[T1]{fontenc}
\usepackage{textcomp}
\usepackage{gensymb}

\usepackage[export]{adjustbox}

\usepackage{algorithm,algorithmicx,algpseudocode}

\usepackage{color}

\usepackage[colorlinks=true,linkcolor=blue,citecolor=blue]{hyperref}

\usepackage{comment}

\usepackage[draft]{todonotes}   % notes shown

\usepackage{soul}

\definecolor{darkgreen}{rgb}{0.53, 0.66, 0.42}
\definecolor{azure}{rgb}{0.0, 0.5, 1.0}

\journal{} 
\usepackage{array}
\newcolumntype{P}[1]{>{\centering\arraybackslash}p{#1}}

\algdef{SE}[SUBALG]{Indent}{EndIndent}{}{\algorithmicend\ }%
\algtext*{Indent}
\algtext*{EndIndent}

\begin{document}

\begin{frontmatter}



%Multi-view Multi-layer
%\title{Benchmarking Graph Neural Networks Based on Reproducibility}
\title{Replica Tree-based Federated Learning using Limited Data}

\author{Ramona Ghilea \fnref{BASIRA}}
\author{Islem Rekik \corref{cor} \fnref{BASIRA}}

\cortext[cor]{Corresponding author: i.rekik@imperial.ac.uk, \url{http://basira-lab.com/}.}
%\cortext[ADNI]{ }

\address[BASIRA]{BASIRA Lab, Imperial-X and Department of Computing, Imperial College London, London, UK}
% \address[Manouba]{National School for Computer Science, University of Manouba, Tunisia}
% \address[DUNDEE]{School of Science and Engineering, Computing, University of Dundee, UK \ }


\begin{abstract}
Learning from limited data has been extensively studied in machine learning, considering that deep neural networks achieve optimal performance when trained using a large amount of samples. Although various strategies have been proposed for centralized training, the topic of federated learning with small datasets remains largely unexplored. Moreover, in realistic scenarios, such as settings where medical institutions are involved, the number of participating clients is also constrained. In this work, we propose a novel federated learning framework, named \emph{RepTreeFL}. At the core of the solution is the concept of a replica, where we replicate each participating client by copying its model architecture and perturbing its local data distribution. Our approach enables learning from limited data and a small number of clients by aggregating a larger number of models with diverse data distributions. Furthermore, we leverage the hierarchical structure of the clients network (both original and virtual), alongside the model diversity across replicas, and introduce a diversity-based tree aggregation, where replicas are combined in a tree-like manner and the aggregation weights are dynamically updated based on the model discrepancy. We evaluated our method on two tasks and two types of data, graph generation and image classification (binary and multi-class), with both homogeneous and heterogeneous model architectures. Experimental results demonstrate the effectiveness and outperformance of \emph{RepTreeFL} in settings where both data and clients are limited. Our code is available at \url{https://github.com/basiralab/RepTreeFL}.
\end{abstract}

%%%%%%%%%%%%%%%%%%%%%%%%%%%%%%%%%%%%

\end{frontmatter}

%\linenumbers

%% ***************************************************************************** %%
\section{Introduction}
%% ***************************************************************************** %%

% Intro limited data in ML + FL, limited data and small no clients challenges in FL
Limited data has been an active area of research in machine learning, given that the performance of deep neural networks is highly dependent on the size of the dataset they are trained on. Different methods have been proposed to address this challenge, such as transfer learning \citep{transfer_learning}, data augmentation \citep{data_augmentation}, and more recently, generating synthetic samples using Generative Adversarial Networks (GANs) \citep{gans}. However, applying these techniques where data resides on different devices is not straightforward. 

Federated Learning \citep{fl_review} is an approach that can be applied in the setting where data is spread across multiple machines, as it aggregates a set of models with different datasets without compromising the data privacy. While various challenges in federated learning have been analysed, such as differences in data distributions across clients \citep{fedprox,feddyn,model_mixing,apfl,hierarchical_clustering,ifca,fedper,fedfomo} or model heterogeneity \citep{fjord,heterofl,federated_dropout,fedrolex,hypernetwork-hetero-fl,graph-hypernetwork-hetero-fl,fedmd,fedgkt}, the limited data issue has not been addressed often. Furthermore, federating a small number of models, which is the case in specific real-world scenarios, is largely unexplored. Therefore, the key research question becomes: \emph{How can we learn efficient models in federated learning in a setting where both data and the participating clients are limited?}

% Proposed method
In this work, we propose a new federated learning method which enables the learning from limited data with only a small number of participating clients, named \emph{RepTreeFL}. In this setting, each client is replicated for a number of times and the original data distribution is modified at each replica, which acts as a virtual copy of the client. The underlying idea is that a greater number of clients with diverse data distributions could enhance the ability of the global model to generalize. However, in real-world scenarios where medical data and healthcare institutions are involved, the number of participating hospitals (clients) is often limited and their local datasets have small sizes. Beyond replicating only the original clients, replicas creation can be done recursively, such that a replica of a replica is created, in order to obtain an even larger amount of diverse models. By leveraging the tree structure of the original and virtual clients and the model diversity across replicas, we introduce a diversity-based tree aggregation, where replicas are aggregated in a tree-like manner and the aggregation weights are dynamically computed based on the discrepancy between models. We evaluate the solution in two different settings: graph generation with heterogeneous models (where architectures vary in dimension) and image classification (binary and multi-class) with homogeneous models. The replica-based federation was published originally in our previous work \citep{repfl}. This paper extends our earlier work with the following contributions:

\begin{enumerate}
    \item We propose a novel diversity-based tree aggregation, which leverages the tree structure of the client network and the model diversity across replicas to enhance the model performance.
    \item We introduce a new hyperparameter to the framework: the tree depth at the client level. In the proposed method, replicas can be created at varying depths, generating a tree of models with multi-length pathways.
    \item Beside the graph super-resolution task, we evaluate both RepFL \citep{repfl} and RepTreeFL on a secondary task: image classification. Therefore, we demonstrate the efficacy of the proposed method in a limited data setting with a small number of participating clients on two tasks and two types of data. Additionally, we evaluate the framework using homogeneous and heterogeneous model architectures.
    
    % \item We propose a novel federated learning approach, \emph{RepTreeFL}, which addresses the realistic scenario where the data is limited and the number of clients is small. In \emph{RepTreeFL}, each client is replicated for a specific number of times and the data distribution on each replica is perturbed. Replicas can be created at multiple depths and a diversity-based tree aggregation is applied in order to enhance the model performance by leveraging the tree structure of the client network and the model diversity across replicas.
    % \item In our experiments, we demonstrate the efficacy of the proposed method in a limited data setting with a small number of participating clients on two tasks and two types of data, image classification and graph generation. Additionally, we evaluate the framework using homogeneous and heterogeneous model architectures.
\end{enumerate}

% %% ***************************************************************************** %%
\section{Related Work}
% %% ***************************************************************************** %% \textbf{

% \textbf{Learning from small datasets.} 
\subsection{Learning from small datasets}

Extensive research has been conducted in machine learning to address the challenge of learning from limited data, as collecting large, high-quality datasets is impractical in general. One technique that was proposed to address this issue is transfer learning \citep{transfer_learning,transfer_learning_review,transfer_learning_review_medical}. This approach aims to enhance learning for a target task by leveraging knowledge acquired from a source task. For instance, a model can be initially trained on a large dataset, then the learned weights can be used as a starting point for a new task which requires a much smaller dataset. Another solution is data augmentation, which overcomes the limitation by modifying the training dataset \citep{data_augmentation,data_augmentation_review_medical}. In the context of images, a straightforward approach to augment the data involves applying simple transformations such as rotation, flipping, or cropping. A more recent technique is data synthesis, which can also be viewed as a form of data augmentation. This method consists in generating synthetic data that has similar characteristics to the original training dataset. These new samples can be created using generative models such as Generative Adversarial Networks (GANs) \citep{gans,gans_review}. However, these approaches face various challenges when applied in the federated learning setting, as data augmentation could only be conducted on the local samples, transfer learning requires a public dataset, while GANs need a considerable amount of samples in order to generate high quality outputs.

% \textbf{Federated Learning.}
\subsection{Federated Learning}

Federated Learning is an approach that enables multiple clients to collaborate and train a global model without sharing their local datasets \citep{fl_review}. In this way, models residing on different devices benefit from training from a larger amount of samples while preserving the data privacy. The classic federated learning algorithm is FedAvg \citep{fedavg}, where clients train their local models for multiple epochs and the server averages their parameters at each round. Later, a wide range of proposed methods addressed different challenges in federated learning. 

One of the main limitations in federated learning is the difference between data distributions across clients. Li et al. proposed FedProx \citep{fedprox}, which introduces a additional term to the local objective and penalizes parameters that are at a considerable distance from the global model. FedDyn \citep{feddyn} adds a similar regularisation term, encouraging parameters of client models to converge to global stationary points. A large category of frameworks that address the data heterogeneity issue is personalized federated learning \citep{personalized-fl,personalized_fl_2}. In this setting, instead of training a single global model, each client generates a personalized model that addresses its own requirements. Model mixing is a popular approach in personalized federated learning \citep{model_mixing}. For instance, APFL \citep{apfl} aims to identify the optimal combination of global and local models using a client-specific parameter. Clustered federated learning was also proposed to create personalized models and consist in grouping similar clients and generating a different model for each group. \citep{hierarchical_clustering} introduces hierarchical clustering, where the server recursively creates two clusters of clients by minimising the cosine similarity between models from the two different groups. In a distinct work \citep{ifca}, where the number of clients is fixed, the clients are assigned a cluster at each round. \citep{fedper} created FedPer, where the global model is split into base and personalized layers. While the base layers undergo training via federated averaging, the personalized layers are trained exclusively utilizing the local dataset. A specific type of personalization was proposed in \citep{fedfomo}, where each client federates with other relevant clients.

Other studies focused on federating models with different architectures, where aggregation becomes more challenging. Partial training of the global model was one of the first proposed ideas \citep{fjord,heterofl} for solving this problem. An example from this category is Federated Dropout \citep{federated_dropout}, which involves setting a subset of the activations to zero in each layer of a client network. Next, only the remaining neurons are trained and transmitted to the server in a federation round. FedRolex \citep{fedrolex} suggested an innovative rolling window extraction of the sub-models for aggregation. The window advances in each round and selects a different part of the global model each time, resulting in a uniform training of the server model parameters. Another set of works developed hypernetworks to generate the personalized model weights. pFedHN \citep{hypernetwork-hetero-fl} was initially created for solving the data heterogeneity limitation, but it was also evaluated on models with heterogeneous architectures. The solution uses such a hypernetworks to create weights for each client individually using client embeddings. Graph hypernetworks \citep{graph-hypernetwork-hetero-fl} were later proposed, which operate on the graph representation of the client architecture. Knowledge distillation \citep{kd} was also applied as a solution for federating heterogeneous architectures. FedMD \citep{fedmd} firstly performs transfer learning, as each client trains its model on a public dataset and then on its private dataset, the during the federation, the knowledge is transferred form server to clients. Another study \citep{fedgkt} introduces group knowledge transfer, where knowledge is transferred bidirectionally between clients and server.

While learning from limited data has been extensively studied in machine learning, federated learning with small datasets has been underexplored. One study \citep{fedmd} considers it as a secondary challenge and applies transfer learning to address it, while another work \citep{feddc} proposes permutations of the local models as a solution. However, while the former requires a pubic dataset, the latter may raise privacy concerns, as models are transferred across clients during federation. Moreover, we are also interested in learning with a few participating clients. We previously developed a replica-based federated learning solution for the limited data setting with a small number of participating clients \citep{repfl}. However, we extend the approach presented in our earlier work by taking into account the diversity across models in the aggregation step and making the depth of the replicas variable across clients.

% %% ***************************************************************************** %%
\section{Method}
% %% ***************************************************************************** %%

\subsection{Problem Formulation}
The goal of federated learning is to train and aggregate $m$ clients, where each client has access to its local dataset $D_{i}$. In RepTreeFL\footnote{\url{https://github.com/basiralab/RepTreeFL}}, there are $m$ original clients and each client is replicated for $r_{i}$ times, where $i$ is the index of the client. The original clients are denoted by \emph{anchors} and the virtual copies of these clients as \emph{replicas}. While the model architecture of the replica is identical to the one on the anchor, the dataset is slightly modified, as the original data distribution is perturbed. The perturbation consist in removing a small percentage $p_{i}$ of samples from the original dataset. We previously described the idea of replicating the original clients in federated learning in \citep{repfl}.

Moreover, we introduce a new concept: a replica can also have its own replicas, creating a tree of anchors and replicas. The depth of each anchor is controlled by hyperparameter $d_{i}$ and each replica of a replica is created in the same way as a replica of an anchor, by copying the model architecture and perturbing the dataset of the parent replica this time. Finally, we federate $m + \sum_{i=1}^{m} (r_{i} + r_{i}^{2} + ... + r_{i}^{d_{i}})$ models.

\begin{figure}[h]
    \centering
    \includegraphics[width=\linewidth]{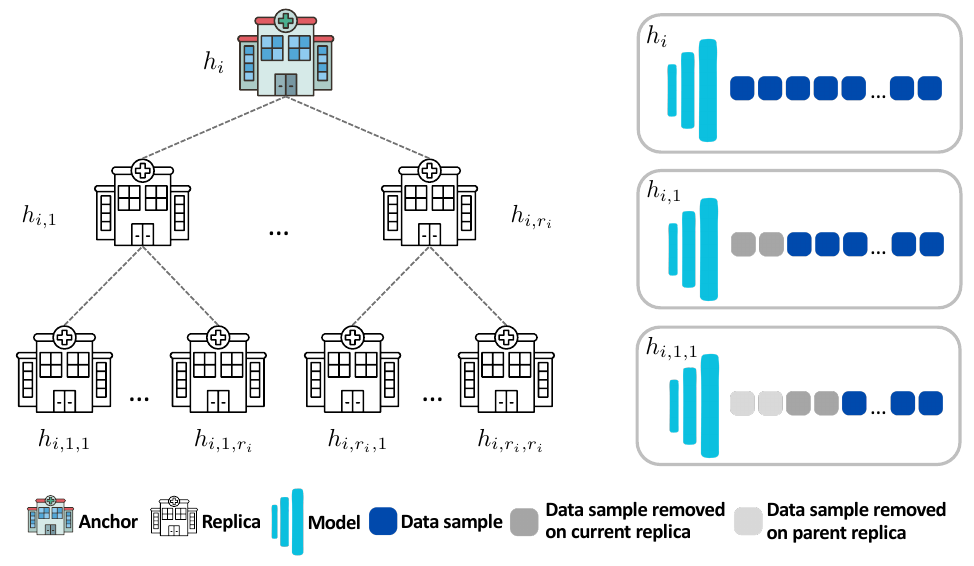}
    \caption{Anchor $i$ and its replicas with $d_{i} = 2$. $h_{i}$ is an anchor hospital (client), $h_{i, 1}$ is its first replica and $h_{i, 1, 1}$ is the first replica of the replica $h_{i, 1}$. At each replica, the model architecture is copied from the parent, while the dataset of the parent is perturbed.}
    \label{fig:main_figure}
\end{figure}

The idea is illustrated in Figure \ref{fig:main_figure}. Client $h_{i}$ is replicated for $r_{i}$ times, then each of its replicas creates its own replicas, resulting in a tree of models. The model architecture is copied at each replica level, while a subset of samples are removed from the parent client.

In our RepTreeFL, the aggregation is performed at each parent node and the aggregation weights are dynamically generated based on a model diversity metric. The local dataset $D_{i}$ of client $i$ contains $n_{i}$ samples. For the classification task, each sample is represented by the image $X_{i, j}$ and the label $y_{i, j}$, where $j$ is the $j$th sample of client $i$. In contrast, for the generation task, each sample is represented by a low-resolution (LR) source graph $X^{s}_{i, j}$ and a high-resolution (HR) target graph $X^{t}_{i, j}$ (with higher resolution than the source graph resolution). Appendix Table \ref{tab:notation_table} summarizes all the mathematical notations used in this work.

\subsection{RepTreeFL: Replica-based Federated Learning}

% \textbf{Replica.} 
\subsubsection{Replica}

At the core of our solution is the concept of a replica, introduced in \citep{repfl}. One limitation that exists in some specific federated learning situations, such as when medical institutions are involved, is the small number of clients that can participate in the federation. The replication of the original clients generates a larger amount of clients, which overcomes this challenge. A second limitation that exists in real-world scenarios is the small amount of data. The solution for this challenge is closely related to the client replicas, as on each virtual client, a slightly different local dataset is created by perturbing the parent (anchor or replica) data distribution. The intuition is that a larger amount of clients with diverse data distributions would enhance the generalisability of the global model. The replica creation is defined by two key aspects:

% \begin{itemize}[wide, labelindent=0pt]
\begin{itemize}
    \item \textbf{Model architecture:} the replica model is identical to the parent model
    \item \textbf{Dataset:} the data distribution of the parent local dataset is perturbed on the replica by removing a small percentage of the samples
\end{itemize}

% \begin{figure}[h]
%     \centering
%     \includegraphics[width=0.6\linewidth]{figures/depth.pdf}
%     \caption{Anchor and replicas with $d_{i} = 2$}
%     \label{fig:anchor_replicas}
% \end{figure}

\begin{figure}[h]
    \centering
    \includegraphics[width=\linewidth]{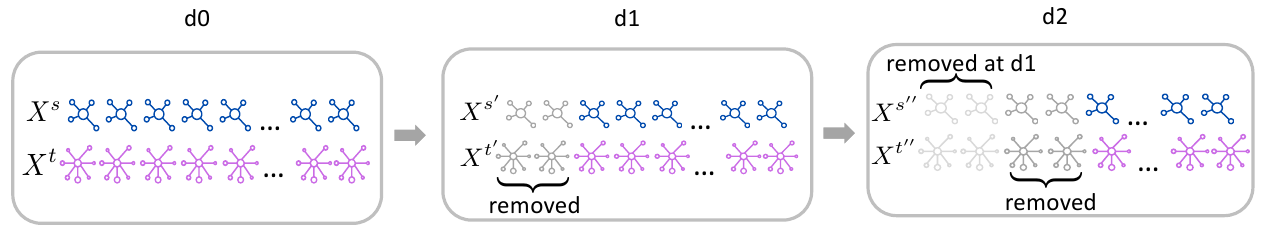}
    \caption{Perturbation of the data distribution at the anchor and replicas with $d_{i} = 2$.}
    \label{fig:per_d2}
\end{figure}

The hierarchical perturbation of the data distribution for the graph generation task is illustrated in Figure \ref{fig:per_d2}. At each replica at level $d = 1$, a small amount of samples is removed from both the set of source graphs $X^{s}$ and the set of target graphs $X^{t}$ at the anchor, resulting in the new dataset consisting of the source graphs $X^{s'}$ and the set of target graphs $X^{t'}$. Furthermore, for each instance of an anchor, the removed samples will vary. For example, if at replica $l$ of client $i$, where the perturbation rate is $p_{i} = 10\%$ and the dataset size is $n_{i} = 20$, samples with indices $j$ and $j + 1$ are excluded, at the subsequent replica $l + 1$ from the same client, the samples with indices $j + 2$ and $j + 3$ will be excluded. If the depth $d_{i}$ of the current anchor is 2, at each replica at level $d = 2$, a small amount of samples is removed from the sets of graphs $X^{s'}$ and $X^{t'}$ at the replica at the previous level, resulting in the new dataset consisting of the set of source graphs $X^{s''}$ and the set of target graphs $X^{t''}$. 

% \begin{figure}[h]
%     \centering
%     \includegraphics[width=\textwidth]{figures/per_d2.pdf}
%     \caption{Perturbation of the data distribution on the replicas with $d_{i} = 2$}
%     \label{fig:per_d2}
% \end{figure}

% \textbf{Diversity-based Aggregation.}
\subsubsection{Diversity-based Aggregation}

We aim to boost the diversity across the generated replicas to enhance the model performance. To achieve this, we propose a model discrepancy metric to quantify the dissimilarity between the anchor model and each of its replicas. Similarly, the metric can be computed if the parent node is a replica itself. The model diversity is computed as follows:
\[div(a, r) = \frac{1}{n_{c}}\sum_{k=1}^{n_{c}} divMetric(a_{k}, r_{k})\]
where $n_{c}$ is the number of aggregated layers, $k$ is the layer index, $a$ is an anchor and $r$ is a replica of anchor $a$. divMetric can be any metric that can quantify the dissimilarity between two layers.

Specifically, we apply the L2 norm to measure the distance between the same layer belonging to two different models. The L2 norm between two layers is computed as follows:
\[||a_{k} - r_{k}||_{2} = \sqrt{\sum_{l=1}^{L}(a_{k}^{l} - r_{k}^{l})^{2}}\]
where $l$ is the index of an element in the weight matrix of the layer $k$ and $L$ is the size of the layer.

Finally, we compute the normalized weights $\alpha$ for replica aggregation based on the diversity metric. In this approach, replica models which are at a larger distance from the anchor model receive greater weights. Given that the anchor is trained on a larger dataset compared to its replicas, we assign it a larger weight for the final anchor model:
\[a \gets \frac{1}{2} (\sum_{r=1}^{r_{i}} \alpha_{r} \times r_{r} + a)\]
where $\alpha_{r}$ is the previously computed weight for replica $r$ and $r_{i}$ is the number of replicas for the current anchor.

The steps taken in the diversity-based tree aggregation are listed in Algorithm \ref{algo:div_tree_agg}.

\begin{algorithm}[H]
\small
% \normalsize
% \scriptsize
\caption{\textit{Diversity-based Tree Aggregation}. $i$ is the anchor index; $r_{i}$ is the number of replicas for anchor $i$; $divArray$ is an array which stores the diversity values between the anchor $i$ and each of its replicas; div is a function which computes the discrepancy between two models;}
\begin{algorithmic}[h]
\State{\textbf{aggregateDiversity($i$):}}
    \Indent
        \For{$r = 1$ \textbf{to} $r_{i}$}
            % \State{$div_{r} \gets ||a_{i} - r||_{2}$}
            \State{$divArray_{r} \gets$ div($a_{i}, r$)}
        \EndFor
        \State{$\alpha \gets$ computeDivAggregationWeights($divArray$)}
        \State{$W_{t + 1, i} \gets \frac{1}{2} (\sum_{r=1}^{r_{i}} \alpha_{r} \times W_{t + 1, r} + W_{t + 1, i}$)}
    \EndIndent

\end{algorithmic}
%\end{scriptize}
\label{algo:div_tree_agg}
\end{algorithm}

% \textbf{RepTreeFL.}
\subsubsection{RepTreeFL}

The pseudocode of \emph{RepTreeFL} is described in Algorithm \ref{algo:replicafl}. At the beginning, each anchor instantiates its replicas, then replicas of replicas are created, according to the depth $d$. Afterwards, at each round, the anchor is trained alongside its virtual copies and the replicas are aggregated hierarchically, at the parent node level, using the dynamically generated weights. Considering that there is a small number of clients, all of them are participating at each round. Finally, the server aggregates the anchors as in classic federated learning.

\begin{algorithm}[H]
\tiny
% \small
% \normalsize
% \footnotesize
% \scriptsize
\caption{\textit{RepTreeFL}. There are $m$ clients indexed by $i$; $E$ is the number of local epochs; $R$ is the number of rounds; $B_{i}$ is the number of local minibatches on client $i$, where the local dataset was split into minibatches of size $B$; $r_{i}$ is the number of replicas for client $i$; $d_{i}$ is the depth for client $i$; $\eta$ is the learning rate;}
\begin{algorithmic}[1]
    \State{\textbf{INPUTS}:\newline \indent $D_{i, j}= \{X_{i, j}, y_{i, j}\}$: image and label of the $j^{th}$ subject in local dataset $D_{i}$ of client $i$}  \newline
    
    \State{\textbf{Server executes:}}
    \Indent
        \For{each client $i = 1$ \textbf{to} $m$}
        \State{createReplicas($i, r_{i}, d_{i}$)}
        \EndFor
    
        \For{each round $t = 1$ \textbf{to} $R$}
        \For{each client $i = 1$ \textbf{to} $m$}
            \State{$W_{t + 1, i} \gets$ clientUpdate($i, W_{t, i}$)}
        \EndFor
        \State{$W_{t + 1} \gets \frac{1}{m} \sum_{i=1}^{m} W_{t + 1, i}$}
        \EndFor
    \EndIndent
    % \State{Send $W^{C}$ to clients} \newline
    \newline

    \State{\textbf{createReplicas($i, repNo, depth$):}}
    \Indent
        \If{$depth > 0$} 
            \For{$r = 1$ \textbf{to} $repNo$}
                % \State{create replica $r$ of client $i$}
                \State{createReplicas($r, repNo, depth - 1$)}
            \EndFor
        \EndIf
    \EndIndent
    \newline
    
    \State{\textbf{clientUpdate($i, W_{t}$):}}
    \Indent
        \For{each local epoch $e = 1$ \textbf{to} $E$}
            \For{each minibatch $b = 1$ \textbf{to} $B_{i}$}
                \State{$W_{t + 1} \gets W_{t} - \eta \bigtriangledown  l(W_{t}, b)$}
                % \State{$W^{D}_{i} \gets W^{D}_{i} - \eta \bigtriangledown  l(W^{D}_{i}, b)$} \Comment{update the personalized layers on client $i$}
            \EndFor
        \EndFor

        \If{hasReplicas($i$)}
            \For{each replica $r = 1$ \textbf{to} $r_{i}$} \Comment{run one round for each replica}
                \State{$W_{t + 1, r} \gets$ clientUpdate($r, W_{t, r}$)}
            \EndFor
            \State{$\alpha \gets$ computeDivAggregationWeights($i$)}
            \State{$W_{t + 1, i} \gets \frac{1}{2} (\sum_{r=1}^{r_{i}} \alpha_{r} \times W_{t + 1, r} + W_{t + 1, i}$)}
        \EndIf
        \State{\Return $W_{t + 1}$}
    \EndIndent
        
    % \State{\textbf{OUTPUTS}: heat-trace vector of subject $i$, $[h^i_{t^1}, h^i_{t^2}, \dots, h^i_{t^{n_t}}]$}
\end{algorithmic}
%\end{scriptize}
\label{algo:replicafl}
\end{algorithm}

\subsection{RepTreeFL with Heterogeneous Models}

There are cases in federated learning where model architectures are different across clients. For instance, clients might have different compute capacities or they might focus on different tasks, such as in multi-tasking. In this case, models are heterogeneous and only a subset of the layers are common across architectures. We adapt the federation in the heterogeneous models setting by federating only the common layers, while training the personalized layers only locally, as in \citep{heterofl}. The local parameter update is described in Algorithm \ref{algo:hetero}. After the training of replicas, the common layers $W^{C}$ are aggregated at the parent node level (i.e., anchor or replica on the level above) and then sent to the server.

\begin{algorithm}[H]
% \tiny
\small
% \normalsize
% \scriptsize
\caption{\textit{Parameter Update for Heterogeneous Models}. $i$ is the client index; $E$ is the number of local epochs; $R$ is the number of rounds; $B_{i}$ is the number of local minibatches on client $i$, where the local dataset was split into minibatches of size $B$; $\eta$ is the learning rate;}
\begin{algorithmic}[1]
% \State{\textbf{parameterUpdateHeterogeneous($i, W_{t}$):}}
%     \Indent
        % \State{\textbf{ClientUpdate($i, W^{C}$):}}
        \For{each local epoch $e = 1$ \textbf{to} $E$}
            \For{each minibatch $b = 1$ \textbf{to} $B_{i}$}
                \State{$W^{C} \gets W^{C} - \eta \bigtriangledown  l(W^{C}, b)$} \Comment{update the common (federated) layers}
                \State{$W^{D}_{i} \gets W^{D}_{i} - \eta \bigtriangledown  l(W^{D}_{i}, b)$} \Comment{update the personalized layers on client $i$}
            \EndFor
        \EndFor
        % \State{Return $W^{C}$ to server}
    % \EndIndent
\end{algorithmic}
%\end{scriptize}
\label{algo:hetero}
\end{algorithm}

% %% ***************************************************************************** %%
\section{Experiments}
% %% ***************************************************************************** %%

\subsection{Datasets and Models}

We evaluate the performance of RepTreeFL in two settings: on image classification with homogeneous models and on graph super-resolution with heterogeneous models.

For the image classification task, two datasets from the MedMNIST benchmark \citep{medmnistv1,medmnistv2} were selected: PneumoniaMNIST for binary classification and OrganCMNIST for multi-class classification \citep{organmnist1,organmnist2}. Given that our method is developed for limited data setting, we evaluate the framework on a subset of the train set, such that each client has a local dataset of 200 samples. As model architecture, we use ResNet10, a smaller variant of the well-known ResNet18 architecture \citep{resnet18}. In this task, all layers are federated, considering that the clients have the same model architecture. 

\begin{table}[h]
\caption{Dataset statistics}
\begin{center} % used the environment to augment the vertical space between the caption and the table
% \footnotesize
% \centering
\resizebox{0.8\linewidth}{!}{%
\begin{tabular}{p{4.5cm} p{3cm} p{1cm} p{1.5cm} p{1cm} p{1.5cm} p{1cm}}
\toprule
\textbf{Dataset} & \textbf{Task (No. Classes)} & \textbf{Size} & \textbf{Selected Subset} & \textbf{Client Dataset}\\
\toprule

SLIM \citep{slim-dataset} & Graph Generation & 279 & 279 & 55\\
\midrule
PneumoniaMNIST \citep{medmnistv1,medmnistv2} & Binary-Class(2) Classification & 4,708 & 800 & 200\\
\midrule
OrganCMNIST \citep{medmnistv1,medmnistv2,organmnist1,organmnist2} & Multi-Class(11) Classification & 13,000 & 800 & 200\\

\bottomrule
\end{tabular} }
\end{center}
\label{tab:datasets}
\end{table}

For the graph super-resolution task, we used a brain graph dataset derived from the Southwest University Longitudinal Imaging Multimodal (SLIM) Brain Data Repository \citep{slim-dataset} using the parcellation approaches described in \citep{parcellation_1} and \citep{parcellation_2}. This dataset contains 279 subjects represented by three brain connectomes: a morphological brain network with resolution 35 (where the resolution denotes the number of brain regions or nodes) derived from T1-weighted MRI, a functional brain network with resolution 160 derived from resting-state functional MRI and a functional brain network with resolution 268 derived from resting-state functional MRI. Further information about the multi-resolution brain graph dataset is available at \citep{slim_dataset_descr}. The models in this setting can have one of two architectures: $G_{1}$, which takes as input a morphological brain connectome of resolution 35 and outputs a functional brain graph of resolution 160, or $G_{2}$, which takes as input a morphological brain connectome of resolution 35 and outputs a functional brain graph of resolution 268. Both models have an graph neural network (GNN) encoder-decoder architecture, which contains edge-based graph convolutional network (GCN) layers \citep{dynamic-edge-graph-conv} and implement the mapping function defined in SG-Net \citep{sgnet}, which creates a symmetric brain graph \footnote{The mapping function and graph preprocessing steps were implemented using code from \url{https://github.com/basiralab/SG-Net}.}. Given that clients have heterogeneous architectures, the parameter update with common and personalized layers described previously is used. The statistics of the datasets are listed in Table \ref{tab:datasets}.

% \begin{table}[h]
% \caption{Dataset statistics}
% \begin{center} % used the environment to augment the vertical space between the caption and the table
% % \footnotesize
% % \centering
% \resizebox{0.9\linewidth}{!}{%
% \begin{tabular}{p{4.5cm} p{3cm} p{1cm} p{1.5cm} p{1cm} p{1.5cm} p{1cm}}
% \toprule
% \textbf{Dataset} & \textbf{Task (No. Classes)} & \textbf{Size} & \textbf{Selected Subset} & \textbf{Client Dataset}\\
% \toprule

% SLIM \citep{slim-dataset} & Graph Generation & 279 & 279 & 55\\
% \midrule
% PneumoniaMNIST \citep{medmnistv1,medmnistv2} & Binary-Class(2) Classification & 4,708 & 800 & 200\\
% \midrule
% OrganCMNIST \citep{medmnistv1,medmnistv2,organmnist1,organmnist2} & Multi-Class(11) Classification & 13,000 & 800 & 200\\

% \bottomrule
% \end{tabular} }
% \end{center}
% \label{tab:datasets}
% \end{table}

\subsection{Baselines}

We compare RepTreeFL against the standalone version (where no federation is applied) and four other federated learning algorithms: FedAvg \citep{fedavg}, FedDyn \citep{feddyn}, FedDC \citep{feddc} and RepFL \citep{repfl}. Details of the baselines are presented below.

\textbf{Standalone} refers to the setting where each client is trained exclusively on its own local dataset and no federation is performed.

\textbf{FedAvg} \citep{fedavg} is the first federated learning algorithm proposed and the most widely applied. In each round, clients train their local models for multiple steps, then communicate their updates to the server, the server computes the average of all received parameters and transfers the new average to the clients.

\textbf{FedDyn} \citep{feddyn} extends FedAvg \citep{fedavg} by introducing a dynamic regularizer to the local objective. It addresses the limitation of statistical heterogeneity across clients while reducing the communication costs.

\textbf{FedDC} \citep{feddc} is a solution specifically developed for federated learning with limited data. The approach consist in transferring models across clients and training them on the other clients local datasets.

\textbf{RepFL} \citep{repfl} is our previously proposed federated learning framework for the limited data setting with a small amount of clients. It is a simpler version of RepTreeFL, where the depth is set to $d = 1$ and a simple aggregation is performed (i.e. all anchors and replicas are assigned the same weights for aggregation).

For the graph super-resolution task, where the models have heterogeneous architectures, FedAvg, FedDyn and FedDC were adapted and applied only on the common layers.

% We compare RepTreeFL against the standalone version (where no federation is applied) and three other federated learning algorithms: FedAvg \citep{fedavg}, FedDyn \citep{feddyn}, which are classic federated learning methods, and FedDC \citep{feddc}, a solution developed for the limited data setting, and RepFL \citep{repfl}, which is a simpler version of the proposed method. In RepFL, the depth is set to $d = 1$ and a simple aggregation is performed, where all anchors and replicas are assigned the same weights. For the graph super-resolution task, where the models have heterogeneous architectures, FedAvg, FedDyn and FedDC were adapted and applied only on the common layers.

\subsection{Experimental Setup}
For image classification, the number of clients is set to three. In the graph generation setting, four clients are used: two clients with the target resolution 160 and two clients with the target resolution 268. The datasets are split between clients in order to simulate the federated learning setting. The framework is evaluated using four fold cross-validation in the image classification setting and five fold cross-validation in the second setting. For each fold configuration, each client receives one fold as its local dataset and the remaining fold is used as a global test set. For both tasks, the number of local epochs is set to $E = 10$ and the number of rounds is $R = 10$. For FedDyn, $\alpha$ is set to 0.01, for FedDC, the daisy-chaining period is 1, and in RepFL, we keep the same perturbation rate and number of replicas as in the proposed method. For image classification, the batch size is 20, the learning rate is set to 0.005 for PneumoniaMNIST \citep{medmnistv1,medmnistv2} and to 0.001 for OrganCMNIST \citep{medmnistv1,medmnistv2,organmnist1,organmnist2}. The optimizer for PneumoniaMNIST is SGD and for OrganCMNIST we use Adam. The loss function applied is cross-entropy loss. For graph generation, a batch size of 5 is used, the learning rate is set to 0.1, SGD is used as optimizer and as loss function, the L1 loss is applied. For simplicity, we fix the number of replicas $r$, the perturbation rate $p$ and the depth $d$ across clients, but they can be variable.

\subsection{Performance comparison to baselines}
\begin{figure}[h]
    \centering
    \includegraphics[width=\linewidth]{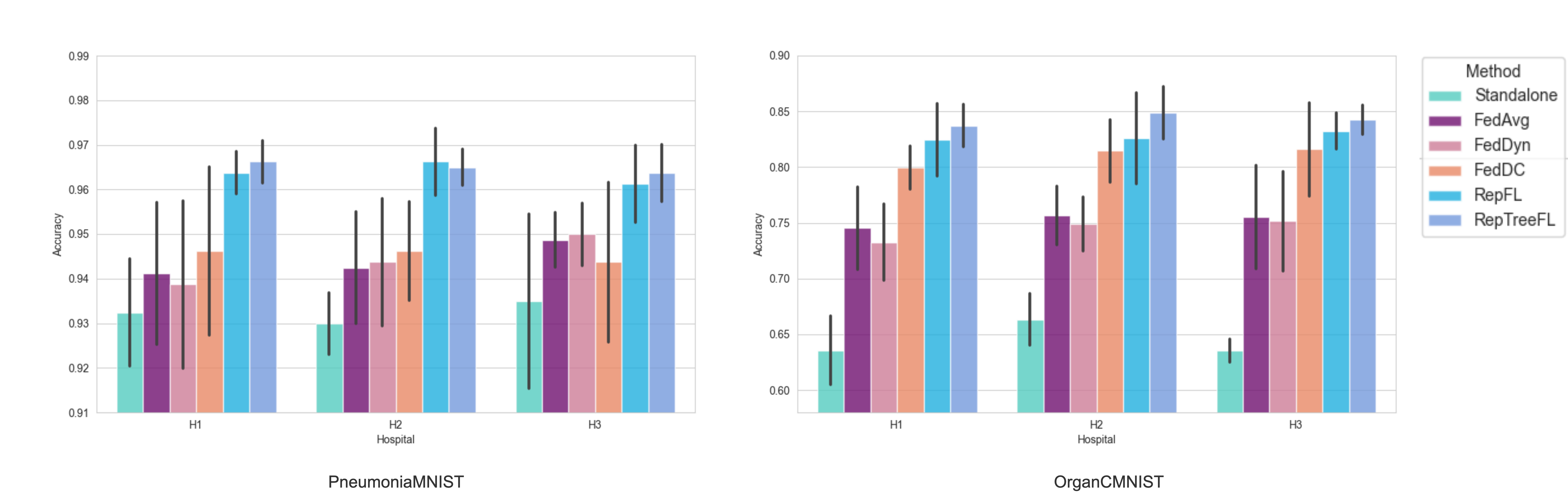}
    \caption{Accuracy on PneumoniaMNIST and OrganCMNIST. H1, H2 and H3 are clients (hospitals) having the same model architecture. Perturbation rate is $p = 10$, number of replicas is $r = 3$ and depth is $d = 1$.}
    \label{fig:medmnist_results}
\end{figure}

\begin{figure}[h]
    \centering
    \includegraphics[width=\linewidth]{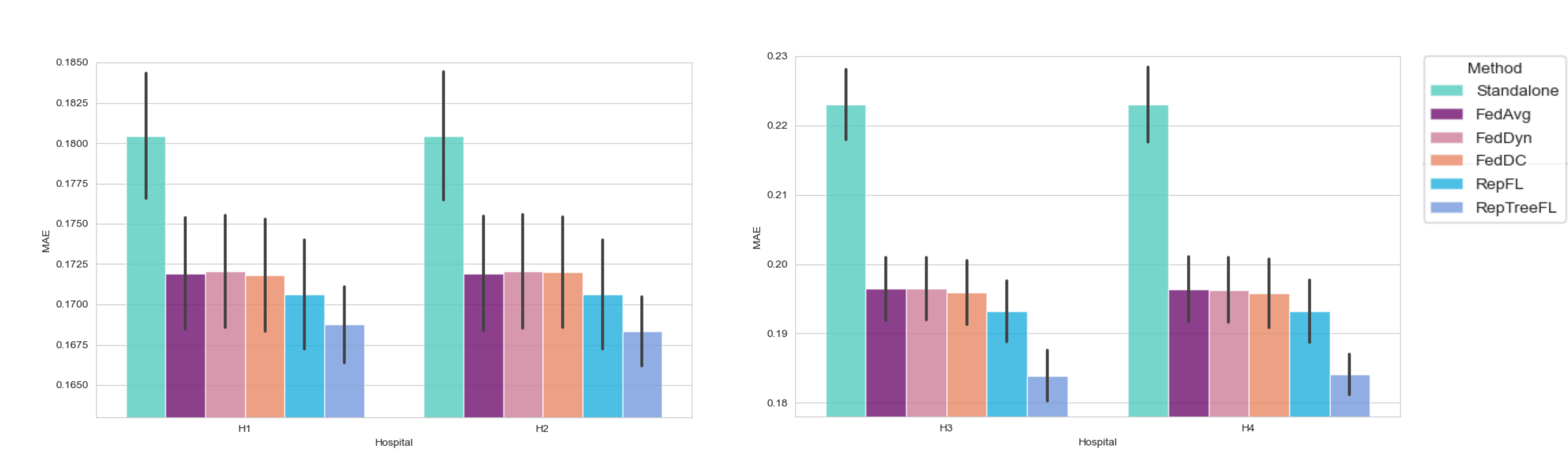}
    \caption{Mean Absolute Error (MAE) on SLIM. Clients H1 and H2 have as super-resolution GNN model $G_{1}$, while clients H3 and H4 have architecture $G_{2}$. Perturbation rate is $p = 10$, number of replicas is $r = 5$ and depth is $d = 1$.}
    \label{fig:slim_results_mae}
\end{figure}

The comparison to baseline methods in the image classification task can be seen in Tables \ref{tab:pneumonia_results} and \ref{tab:organ_results} and in the graph super-resolution task in Table \ref{tab:slim_results}. We report the accuracy, F1 score, sensitivity and specificity on the image classification task, and the mean absolute error (MAE) in the graph generation task. The values are reported for each client model on the global test set across four and five folds, respectively. Additionally, the accuracy comparison in image classification is shown in Figure \ref{fig:medmnist_results} and the MAE in graph generation is illustrated in Figure \ref{fig:slim_results_mae}.

First, the federated learning algorithms show significant performance gain compared to the standalone version on both tasks and on all three datasets. These results demonstrate that federated learning enables the clients to efficiently collaborate and benefit from complementary knowledge, which is especially important when the local datasets are small.

Second, the proposed method, RepTreeFL, generally outperforms all baselines in both settings, image classification (both binary and multi-class) with homogeneous models and graph generation with heterogeneous models. The only exception is hospital (client) H2 on the PneumoniaMNIST \citep{medmnistv1,medmnistv2} dataset, where RepFL \citep{repfl}, our previously developed method, reaches slightly better results across all metrics. Moreover, RepTreeFL achieves the smallest standard deviation across folds on both tasks and all datasets.

\begin{table}[h]
\caption{Comparison to baselines on PneumoniaMNIST}
\centering
\resizebox{\linewidth}{!}{%
\begin{tabular}{@{}ccccccccccccc@{}}
\toprule
\multirow{2}{*}{\textbf{Method}} &
  \multicolumn{4}{c}{\textbf{H1}} &
  \multicolumn{4}{c}{\textbf{H2}} &
  \multicolumn{4}{c}{\textbf{H3}} \\

\cmidrule(l){2-13}

& \textbf{Acc} & \textbf{F1} & \textbf{Sens} & \textbf{Spec} & \textbf{Acc} & \textbf{F1} & \textbf{Sens} & \textbf{Spec} &  \textbf{Acc} & \textbf{F1} & \textbf{Sens} & \textbf{Spec} \\          

\midrule

\textbf{Standalone} 
& $93.24$ & $91.18$ & $91.34$ & $91.34$ &
$92.99$ & $90.80$ & $90.69$ & $90.69$ &  
$93.49$ & $91.42$ & $91.20$ & $91.20$ \\ \midrule

% \textbf{\shortstack{FedAvg \\ \citep{fedavg}}} 
\textbf{FedAvg} 
& $94.11$ & $92.31$ & $92.40$ & $92.40$ &
$94.24$ & $94.24$ & $94.24$ & $92.80$ &  
$94.86$ & $93.28$ & $93.22$ & $93.22$ \\ \midrule

\textbf{FedDyn} 
& $93.87$ & $91.96$ & $92.08$ & $92.08$ &
$94.37$ & $92.70$ & $92.88$ & $92.88$ &  
$94.99$ & $93.46$ & $93.46$ & $93.46$ \\ \midrule

\textbf{FedDC} 
& $94.62$ & $92.94$ & $92.73$ & $92.73$ &
$94.62$ & $93.05$ & $93.66$ & $93.66$ &  
$94.37$ & $92.70$ & $93.03$ & $93.03$ \\ \midrule

\textbf{RepFL (ours)} 
& $96.37$ & $95.22$ & $94.87$ & $94.87$ &
$\mathbf{96.62}$ & $\mathbf{95.57}$ & $\mathbf{95.51}$ & $\mathbf{95.51}$ &  
$96.12$ & $94.85$ & $94.22$ & $94.22$ \\ \midrule

\textbf{RepTreeFL (ours)} 
& $\mathbf{96.62}$ & $\mathbf{95.56}$ & $\mathbf{95.36}$ & $\mathbf{95.36}$ &
$96.50$ & $95.40$ & $95.27$ & $95.27$ &  
$\mathbf{96.37}$ & $\mathbf{95.18}$ & $\mathbf{94.54}$ & $\mathbf{94.54}$ \\

\bottomrule

\end{tabular}
}
\label{tab:pneumonia_results}
\end{table}

\begin{table}[h]
\caption{Comparison to baselines on OrganCMNIST}
\centering
\resizebox{\linewidth}{!}{%
\begin{tabular}{@{}ccccccccccccc@{}}
\toprule
\multirow{2}{*}{\textbf{Method}} &
  \multicolumn{4}{c}{\textbf{H1}} &
  \multicolumn{4}{c}{\textbf{H2}} &
  \multicolumn{4}{c}{\textbf{H3}} \\

\cmidrule(l){2-13}

& \textbf{Acc} & \textbf{F1} & \textbf{Sens} & \textbf{Spec} & \textbf{Acc} & \textbf{F1} & \textbf{Sens} & \textbf{Spec} &  \textbf{Acc} & \textbf{F1} & \textbf{Sens} & \textbf{Spec} \\          

\midrule

\textbf{Standalone} 
& $63.56$ & $58.28$ & $57.85$ & $96.19$ &
$66.33$ & $63.21$ & $62.65$ & $96.48$ &  
$63.56$ & $59.23$ & $58.17$ & $96.19$ \\ \midrule

% \textbf{\shortstack{FedAvg \\ \citep{fedavg}}} 
\textbf{FedAvg} 
& 
$74.53$ & $71.66$ & $70.64$ & $97.35$ &
$75.66$ & $73.72$ & $72.08$ & $97.46$ &  
$75.54$ & $73.14$ & $72.36$ & $97.48$ \\ \midrule

\textbf{FedDyn} 
& 
$73.27$ & $70.52$ & $69.39$ & $97.22$ &
$74.91$ & $72.78$ & $71.21$ & $97.38$ &  
$75.16$ & $72.94$ & $72.12$ & $97.45$ \\ \midrule

\textbf{FedDC} 
& 
$79.95$ & $78.48$ & $77.61$ & $97.93$ &
$81.46$ & $79.52$ & $79.26$ & $98.10$ &  
$81.59$ & $80.26$ & $79.78$ & $98.10$ \\ \midrule

\textbf{RepFL (ours)} 
& 
$82.47$ & $79.58$ & $80.13$ & $98.24$ &
$82.59$ & $80.12$ & $79.94$ & $98.22$ &  
$83.23$ & $80.90$ & $80.79$ & $98.31$ \\ \midrule

\textbf{RepTreeFL (ours)} 
& 
$\mathbf{83.73}$ & $\mathbf{81.14}$ & $\mathbf{81.31}$ & $\mathbf{98.36}$ &
$\mathbf{84.86}$ & $\mathbf{82.61}$ & $\mathbf{82.83}$ & $\mathbf{98.47}$ &
$\mathbf{84.24}$ & $\mathbf{82.17}$ & $\mathbf{82.06}$ & $\mathbf{98.41}$ \\

\bottomrule

\end{tabular}
}
\label{tab:organ_results}
\end{table}

\begin{table}[h]
\caption{Comparison to baselines on SLIM}
\centering
\resizebox{0.9\linewidth}{!}{%
\begin{tabular}{@{}ccccc@{}}
\toprule
\textbf{Method} &
  \textbf{H1} &
  \textbf{H2} &
  \textbf{H3} &
  \textbf{H4} \\     

\midrule

\textbf{Standalone} & $0.1804 \pm 0.0038$ & $0.1804 \pm 0.0039$ & $0.2230 \pm 0.0051$ & $0.2229 \pm 0.0053$ \\ \midrule

\textbf{FedAvg} & $0.1719 \pm 0.0034$ & $0.1719 \pm 0.0035$ & $0.1964 \pm 0.0045$ & $0.1963 \pm 0.0046$ \\ \midrule

\textbf{FedDyn} & $0.1720 \pm 0.0034$ & $0.1720 \pm 0.0035$ & $0.1964 \pm 0.0045$ & $0.1962 \pm 0.0046$ \\ \midrule

\textbf{FedDC} & $0.1718 \pm 0.0034$ & $0.1720 \pm 0.0034$ & $0.1959 \pm 0.0046$ & $0.1957 \pm 0.0049$ \\ \midrule

\textbf{RepFL (ours)} & $0.1706 \pm 0.0033$ & $0.1706 \pm 0.0033$ & $0.1931 \pm 0.0043$ & $0.1931 \pm 0.0041$ \\ \midrule

\textbf{RepTreeFL (ours)} & $\mathbf{0.1687 \pm 0.0024}$ & $\mathbf{0.1684 \pm 0.0021}$ & $\mathbf{0.1839 \pm 0.0036}$ & $\mathbf{0.1841 \pm 0.0029}$ \\

\bottomrule

\end{tabular}
}
\label{tab:slim_results}
\end{table}

% \begin{figure}[h]
%      \centering
%      \begin{subfigure}[b]{0.49\linewidth}
%          \centering
%          \includegraphics[width=\linewidth]{figures/average_runs_acc.png}
%          \caption{Accuracy comparison to baselines on PneumoniaMNIST}
%         \label{fig:pneumonia_results_acc}
%      \end{subfigure}
%      \hfill
%      \begin{subfigure}[b]{0.49\linewidth}
%          \centering
%          \includegraphics[width=\linewidth]{figures/average_runs_acc_organ.png}
%          \caption{Accuracy comparison to baselines on OrganCMNIST}
%         \label{fig:organ_results_acc}
%      \end{subfigure}
%         \caption{Results on MedMNIST. Perturbation rate is $p = 10$, number of replicas is $r = 3$ and depth is $d = 1$.}
%         \label{fig:medmnist}
% \end{figure}

\subsection{Analysis of the perturbation rate, depth and the diversity-based aggregation}

\begin{figure}[h]
    \centering
    \includegraphics[width=\linewidth]{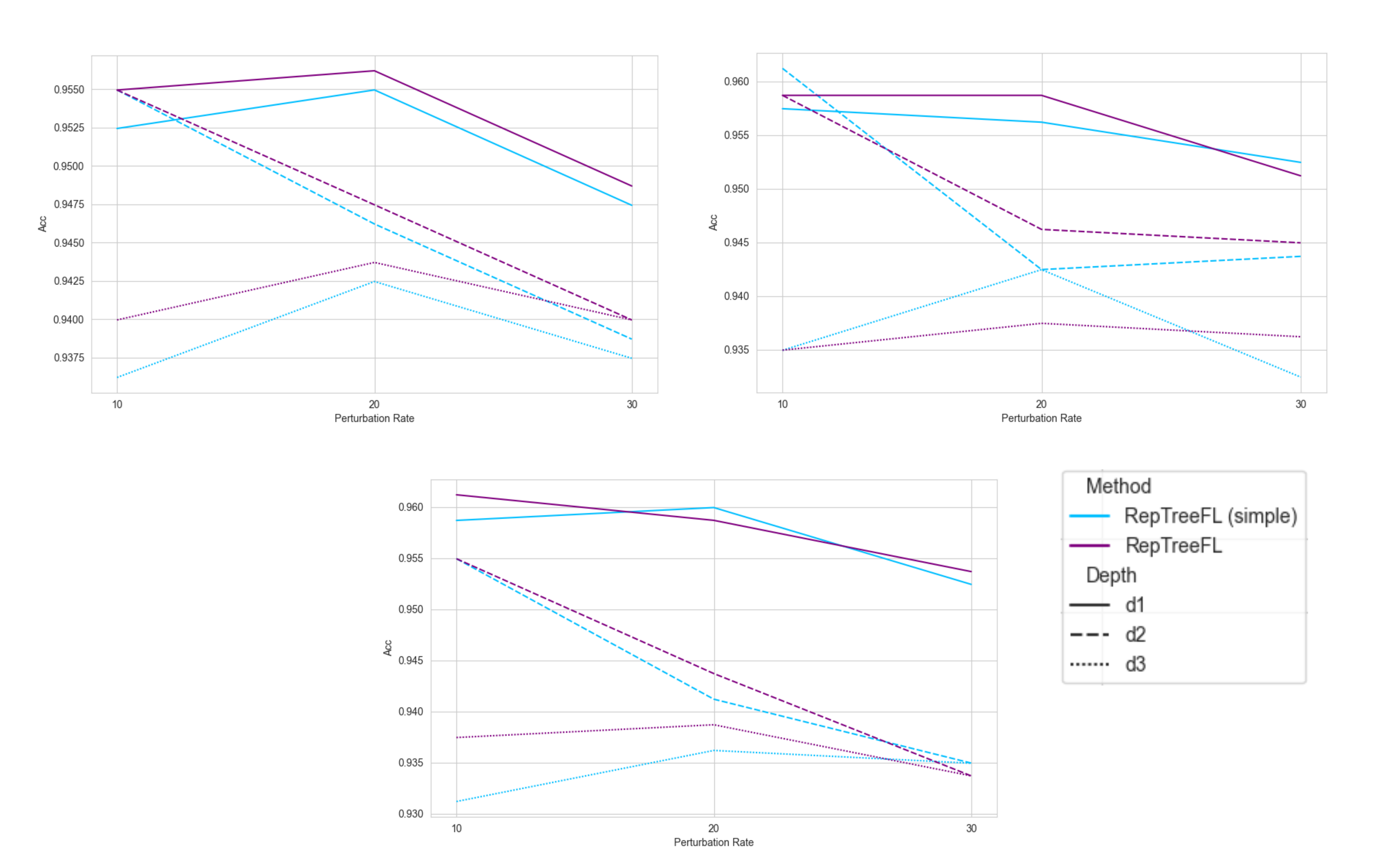}
    \caption{Accuracy analysis of the the rate of perturbation and depth on PneumoniaMNIST. Number of replicas is $r = 3$.}
    \label{fig:depth_per}
\end{figure}

Additionally, we study the influence of two hyperparameters on the framework performance, the rate of perturbation $p$ and the depth $d$. Also, we compare our method, which applies diversity-based aggregation, with simple aggregation, where the replica federation is performed with fixed weights and all clients (original and virtual) are assigned the same weight when aggregated. We analyse the values $p \in \{10\%, 20\%, 30\%\}$ and $d \in \{1, 2, 3\}$. We fix the number of replicas to $r = 3$. We conduct this experiment on PneumoniaMNIST \citep{medmnistv1,medmnistv2}. Figure \ref{fig:depth_per} illustrates our comparative analysis. The simple aggregation version with $d = 1$ is equivalent to RepFL \citep{repfl}. 

Firstly, while the version with simple aggregation, depth $d = 2$ and perturbation rate $p = 10\%$ outperforms the other variants for one client, the difference in accuracy compared to RepTreeFL with depth $d = 1$ is negligible. Therefore, in this setting, the ideal depth is $d = 1$, considering that another level of replicas implies additional computation and memory costs. 

Secondly, the diversity-based tree aggregation version outperforms the simple aggregation in most of the cases, which proves the efficacy of the proposed aggregation approach.

\subsection{Analysis of the perturbation approach}

\begin{figure}[h]
    \centering
    \includegraphics[width=0.6\linewidth]{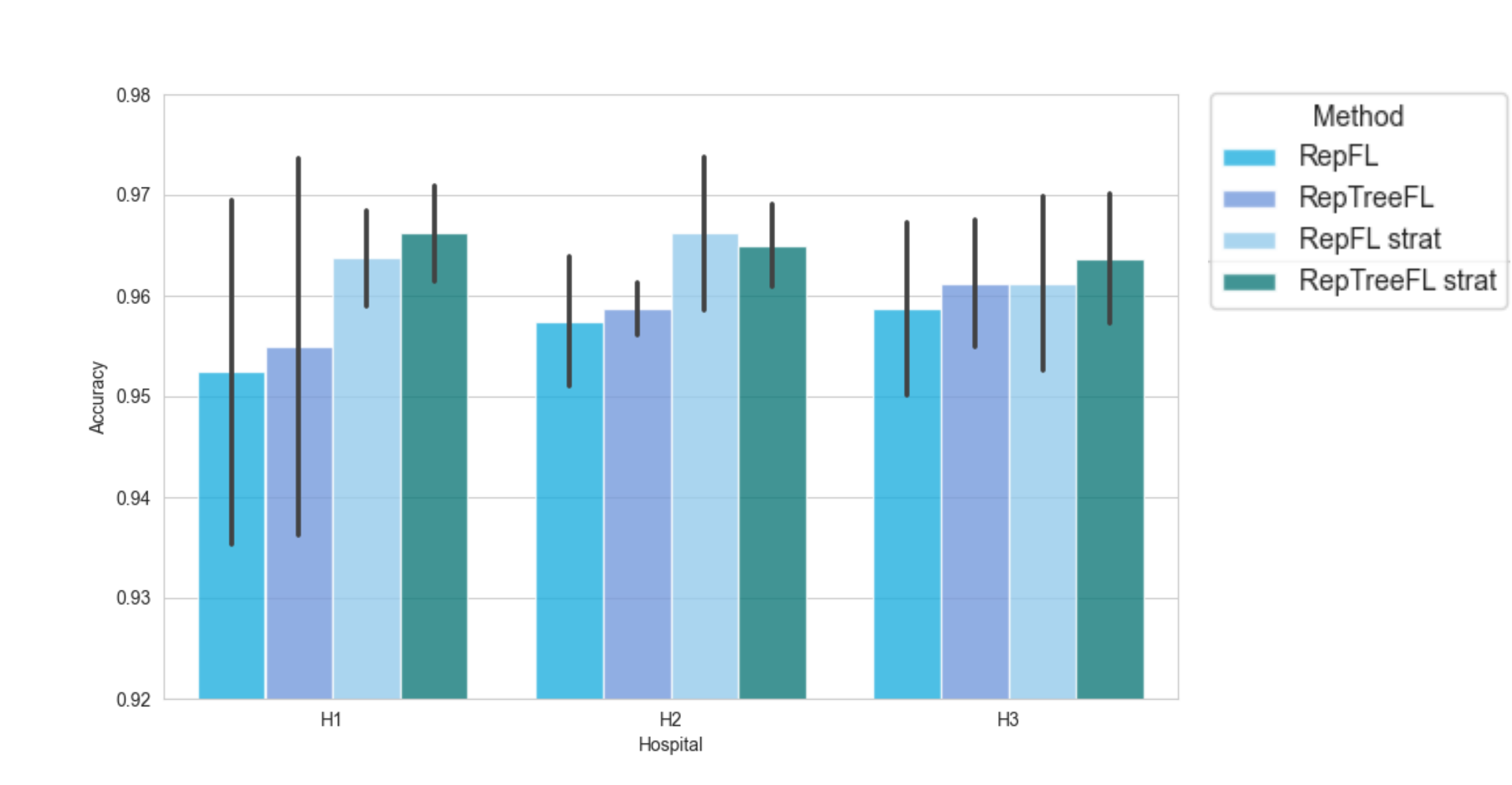}
    \caption{Accuracy analysis of the perturbation approach on PneumoniaMNIST. Perturbation rate is $p = 10\%$, number of replicas is $r = 3$ and depth is $d = 1$.}
    \label{fig:pneumonia_strat}
\end{figure}

Next, we analyse the perturbation approach on PneumoniaMNIST \citep{medmnistv1,medmnistv2} and compare the stratified perturbation to random perturbation on RepFL \citep{repfl} and the proposed method. In stratified perturbation, when removing a subset of samples on each replica, we maintain the same label distribution as in the anchor dataset. We fix the rate of perturbation to $p = 10\%$, the number of replicas to $r = 3$ and the depth to $d = 1$. Figure \ref{fig:pneumonia_strat} shows that RepTreeFL achieves better results when the stratified approach is applied. Moreover, RepFL (which involves simple aggregation) with stratified perturbation outperforms RepTreeFL with random perturbation, which proves the importance of perturbing the local datasets by preserving the label distribution of the anchor dataset.

\subsection{Performance on smaller datasets}

\begin{figure}[h]
    \centering
    \includegraphics[width=\linewidth]{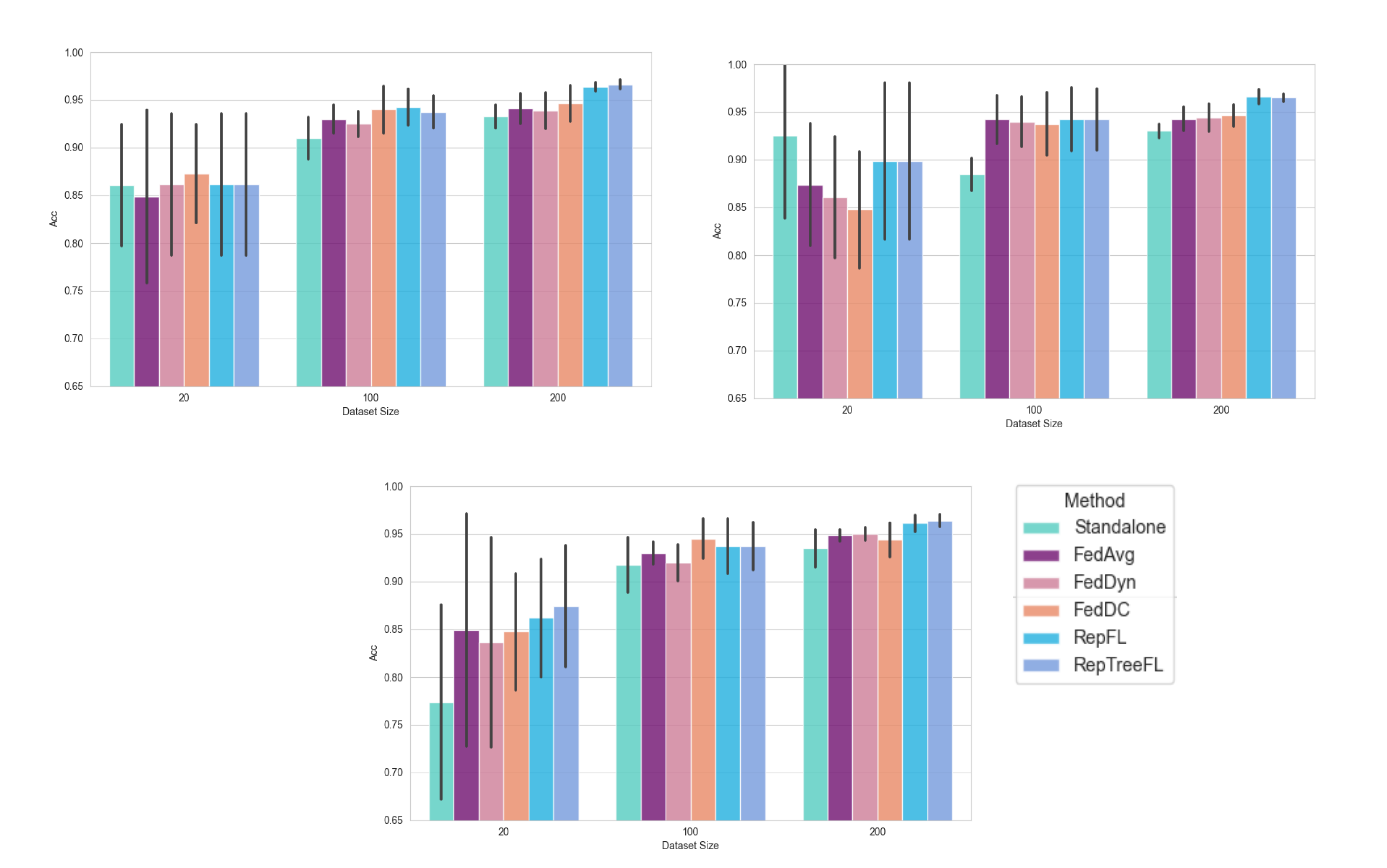}
    \caption{Accuracy analysis of different small dataset sizes on PneumoniaMNIST. Perturbation rate is $p = 10\%$, number of replicas is $r = 3$ and depth is $d = 1$.}
    \label{fig:pneumonia_dataset_size}
\end{figure}

As our solution is developed for a setting where data is limited, we study the behaviour of RepTreeFL on smaller subsets of PneumoniaMNIST \citep{medmnistv1,medmnistv2} and compare it to the baselines. We analyse the dataset size values $n \in \{20, 100, 200\}$. The rate of perturbation is set to $p = 10\%$, to ensure that very small local datasets do not suffer from a significant reduction in the number of removed samples. We also fix the number of replicas to $r = 3$ and the depth to $d = 1$. It can be observed in Figure \ref{fig:pneumonia_dataset_size} that our method outperforms on average the other federated learning methods on local datasets of various small sizes.

\subsection{Comparison to centralized training}

\begin{figure}[h]
    \centering
    \includegraphics[width=\textwidth]{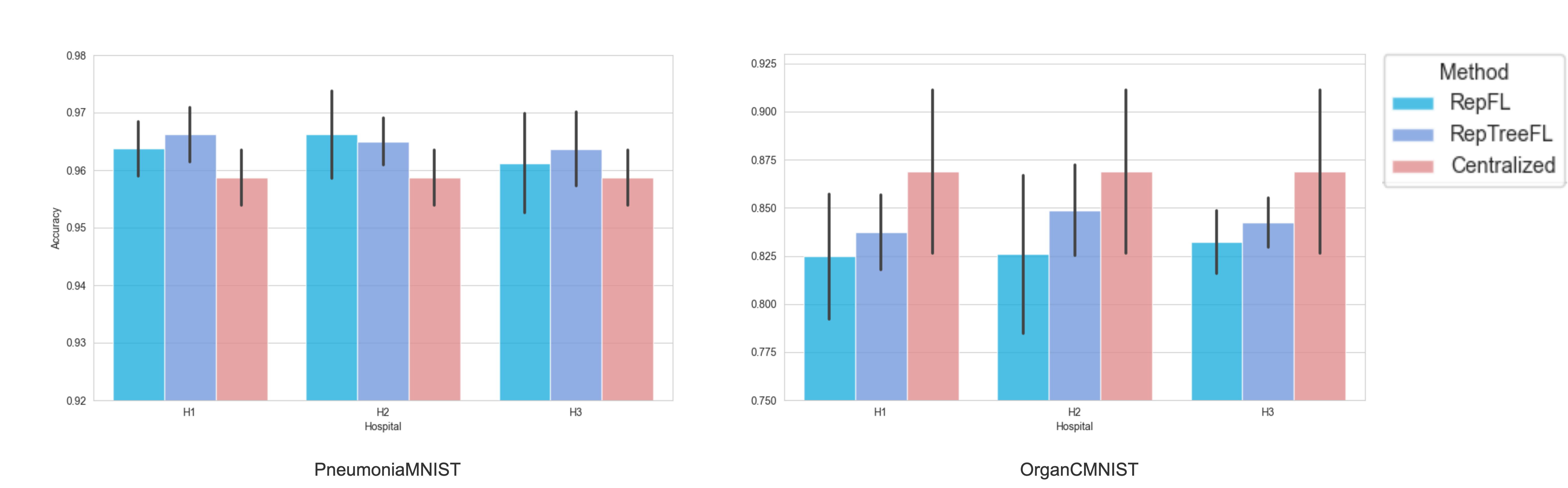}
    \caption{Accuracy comparison to centralized training on MedMNIST}
    \label{fig:medmnist_centralized}
\end{figure}

\begin{figure}[h]
    \centering
    \includegraphics[width=\textwidth]{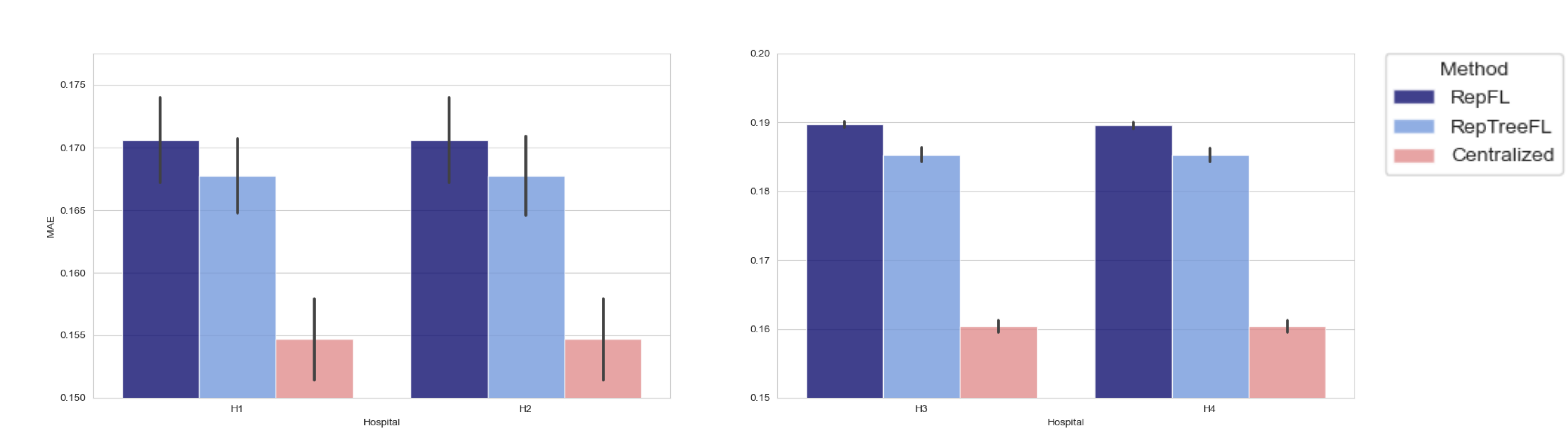}
    \caption{MAE comparison to centralized training. Clients H1 and H2 have as model $G_{1}$, while H3 and H4 have architecture $G_{2}$. The number of replicas is set to $r = 5$.}
    \label{fig:slim_centralized}
\end{figure}

We compare the proposed method to the centralized training setting, where a model is trained using all samples, which were distributed across clients in the federated learning setting. The evaluation is performed using four fold cross-validation for image classification and five fold cross-validation for graph generation, as in the previous experiments. For each configuration in the centralized training setting, three (or four) folds are used for training, and the remaining fold is used for testing. The results for image classification and graph generation can be observed in Figures \ref{fig:slim_centralized} and \ref{fig:medmnist_centralized}, respectively.

Although RepFL \citep{repfl} and RepTreeFL demonstrate accuracy levels similar to centralized training on OrganCMNIST \citep{medmnistv1,medmnistv2,organmnist1,organmnist2}, they surpass centralized training when applied to PneumoniaMNIST \citep{medmnistv1,medmnistv2}. A better performance on PneumoniaMNIST than on OrganCMNIST is expected, as the binary classification task in PneumoniaMNIST is considered easier compared to the multi-classification task in OrganCMNIST. The gain in accuracy of RepTreeFL over centralized training can be explained by the boosting effect of our proposed method, as the framework can be seen as related to ensemble methods.

When analysing the results on the graph generation task, there is a significant difference in MAE between our method and centralized training, as this task involves a higher level of complexity.

\subsection{Computation and memory concerns}

While replicas facilitate learning with limited data and a small amount of clients in federated learning, they are essentially duplicates of the original models and are stored at the anchor clients. This implies extra memory and computational resources, which needs to be considered in future work. One approach to minimise this concern is opting for a model architecture that is as simple as possible.

% %% ***************************************************************************** %%
\section{Conclusion}
% %% ***************************************************************************** %%

We proposed RepTreeFL, a novel framework for the limited data setting, where a small number of clients are participating in the federation process. The solution consists in replicating each original client and perturbing its local dataset, such that the model performance is enhanced by a larger amount of diverse clients. The method combines the tree structure of the client network, as well as the diversity across created replicas, and introduces a diversity-based aggregation approach. Empirical evaluation shows the efficacy of the proposed method on two different settings: image classification with homogeneous models and graph generation with heterogeneous models. Additionally, we provide an analysis of the main hyperparameters, on the perturbation approach, as well as on datasets of variable sizes. Future work includes analysing more complex diversity metrics for aggregation, as well as evaluating the method on non-IID data and decreasing the compute and memory budgets.

\section*{Data Availability}
The data used for the graph super-resolution task is confidential, while the data used for image classification is publicly available at \url{https://medmnist.com/}.

\section*{Code Availability}
The implementation is available at \newline
\url{https://github.com/basiralab/RepTreeFL}.

\section*{Competing Interests}
The authors declare no competing interests.

% \section{Acknowledgements}

%%%%%%%%%%%%%%%%%%%%%%%%%%%%%%%%%%%%%%%%%%%%%%%%%%%%%%%%%%%%%%%%%%%%%%%%%%%%%%%%%%%%%%%%%%%%%%%%%%%%%%%%%%%%
\newpage
\bibliography{Biblio3}
\bibliographystyle{model2-names}

%%%%%%%%%%%%%%%%%%%%%%%%%%%%%%%%%%%%
\newpage
\appendix
% \section{Appendix}

%%%%%%%%%%%%%%%%%%%%%%%%%%%%%%%%%%%%
% Notation Table
\section{Mathematical Notations}

\begin{table}[H]
\caption{Table of mathematical notations}
% \small
% \tiny
\scriptsize
\begin{center}% used the environment to augment the vertical space between the caption and the table
% \centering
\begin{tabular}{c c p{8cm} }
\toprule
Notation & Dimension & Description\\
\toprule
% \multicolumn{3}{c}{}\\
% \multicolumn{3}{c}{\underline{General}}\\
% \multicolumn{3}{c}{}\\

$m$ & 1 & Number of original clients\\
$i$ & 1 & Indicator for client\\
$D_{i}$ & 1 & Local dataset of client $i$\\
$n_{i}$ & 1 & Length of the dataset of client $i$\\

\midrule
% \multicolumn{3}{c}{}\\
% \multicolumn{3}{c}{\underline{Replicas}}\\
% \multicolumn{3}{c}{}\\

$r_{i}$ & 1 & Number of replicas of client $i$\\
$p_{i}$ & 1 & Perturbation rate (percentage) of client $i$\\
$d_{i}$ & 1 & Maximum replica depth of client $i$\\
$j$ & 1 & Indicator for sample\\
$X_{i, j}$ & $\mathbb{R}^{2}$ & Image $j$ of client $i$\\
$y_{i, j}$ & 1 & Label of sample $j$ of client $i$\\
$X^{s}_{i, j}$ & $\mathbb{R}^{2}$ & Source brain graph from sample $j$ of client $i$\\
$X^{t}_{i, j}$ & $\mathbb{R}^{2}$ & Target brain graph from sample $j$ of client $i$\\
$X^{s}$ & $\mathbb{R}^{2}$ & Set of source brain graphs on anchor\\
$X^{t}$ & $\mathbb{R}^{2}$ & Set of target brain graphs on anchor\\
$X^{s'}$ & $\mathbb{R}^{2}$ & Set of source brain graphs on replica at $d = 1$\\
$X^{t'}$ & $\mathbb{R}^{2}$ & Set of target brain graphs on replica at $d = 1$\\
$X^{s''}$ & $\mathbb{R}^{2}$ & Set of source brain graphs on replica at $d = 2$\\
$X^{t''}$ & $\mathbb{R}^{2}$ & Set of target brain graphs on replica at $d = 2$\\
$a$ & 1 & Anchor index\\
$r$ & 1 & Replica index\\
$\alpha$ & $\mathbb{R}$ & Aggregation weights\\

\midrule
% \multicolumn{3}{c}{}\\
% \multicolumn{3}{c}{\underline{Models}}\\
% \multicolumn{3}{c}{}\\

$G_{1}$ & - & Generator 1\\
$G_{2}$ & - & Generator 2\\
$n_{c}$ & 1 & Number of common layers between two heterogeneous models\\
$W^{C}$ & $n_{c} \times \mathbb{R}^{2}$ & Weight matrices of common layers between two heterogeneous models\\

% $TC$ & $1$ & Total overall cost(\$)\\  
% \multicolumn{3}{c}{}\\
% \multicolumn{3}{c}{\underline{Decision Variables}}\\
% \multicolumn{3}{c}{}\\
% $y_f$ & $=$ & \(\left\{\begin{array}{rl}
% 1,  & \text{if Supplier located at site $f$ is open} \\
% 0,  & \text{otherwise} \end{array} \right.\)\\
\bottomrule
\end{tabular}
\end{center}
\label{tab:notation_table}
\end{table}

\end{document}